\documentclass[10pt,twocolumn,letterpaper]{article}

\usepackage{iccv}              

\usepackage{algorithm}
\usepackage{algpseudocode}

\usepackage{multirow}

\definecolor{iccvblue}{rgb}{0.21,0.49,0.74}
\usepackage[pagebackref,breaklinks,colorlinks,citecolor=iccvblue]{hyperref}

\def\paperID{7357} 
\def\confName{ICCV}
\def\confYear{2025}

\title{Multi-Cue Adaptive Visual Token Pruning for Large Vision-Language Models}

\author{Bozhi~Luan$^{1}$~~~~~Wengang~Zhou$^{1}$~~~~~Hao~Feng$^{1}$~~~~~Zhe~Wang$^{2}$~~~~~Xiaosong~Li$^{2}$~~~~~Houqiang~Li$^{1}$\\
{\normalsize $^{1}$ University of Science and Technology of China},
{\normalsize $^{2}$ Huawei Technologies}\\
{\tt\small \{bzluan,haof\}@mail.ustc.edu.cn, \{zhwg,lihq\}@ustc.edu.cn,} \
\tt\small \{wangzhe226,lixiaosong20\}@huawei.com}

\begin{document}
\maketitle

\begin{abstract}
As the computational needs of Large Vision-Language Models (LVLMs) increase, visual token pruning has proven effective in improving inference speed and memory efficiency. Traditional pruning methods in LVLMs predominantly focus on attention scores to determine token relevance, overlooking critical aspects such as spatial position and token similarity. To this end, we introduce AdaptPrune, a novel plug-and-play training-free pruning method that builds on conventional attention-based pruning by integrating spatial distance and token similarity with an adaptive NMS approach. Our method is based on several observed phenomena in large models: the positional bias in the model's image attention and the redundancy of token information ignored by previous approaches. By integrating attention, spatial, and similarity information, our approach ensures a comprehensive evaluation of token importance and substantially refines the pruning decisions. Our method has been extensively tested across various LVLMs and benchmarks, confirming its robustness and adaptability. The results demonstrate that AdaptPrune consistently outperforms existing methods across various pruning ratios. Code is available at \url{https://github.com/bzluan/AdaptPrune}.


\end{abstract}

\begin{figure}[t]
  \centering
  \hspace{-0.8cm} 
  \includegraphics[width=\linewidth]{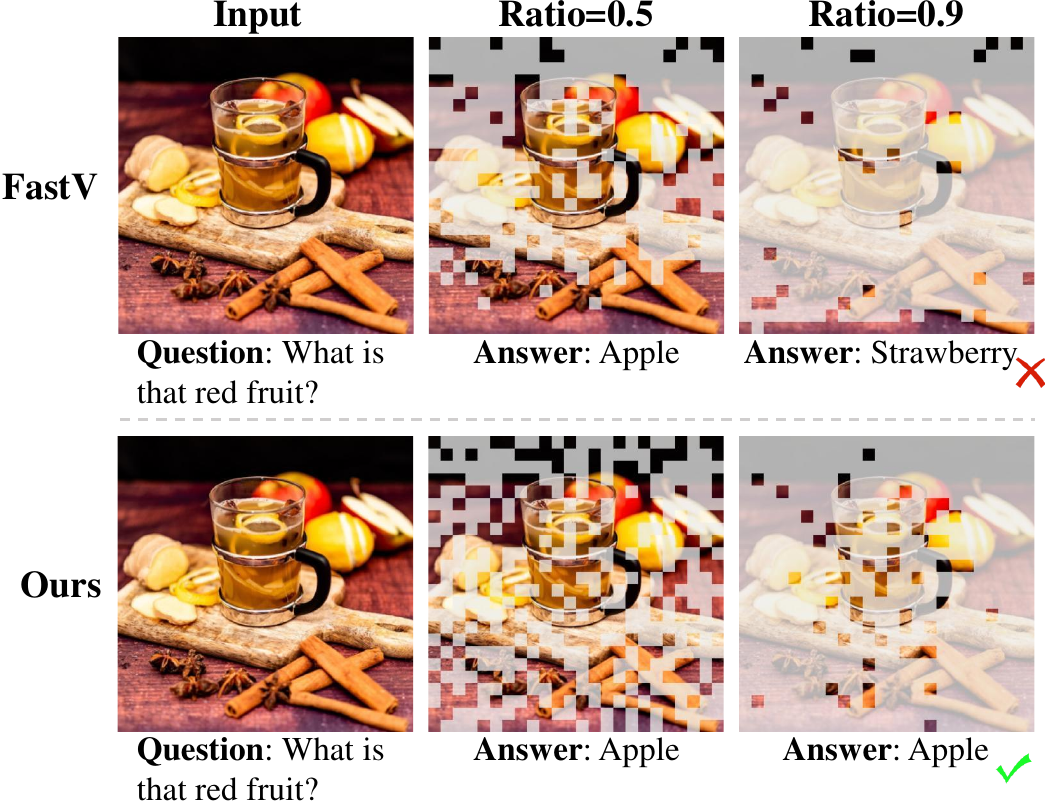}
  \hspace{-0.8cm} 
  \caption{
     Comparison between FastV \cite{fastv} and our AdaptPrune at different pruning ratios. FastV \cite{fastv} utilizes only attention scores for token pruning, while AdaptPrune incorporates attention, spatial, and similarity information for a more holistic investigation.
  }
  \label{fig:figure1}
  \vspace{-8pt}
\end{figure}

\section{Introduction}

Breakthroughs in Large Language Models (LLMs) \cite{llama, chiang2023vicuna, qwen} have revolutionized natural language processing with impressive performance across a diverse range of tasks \cite{wang2018glue,mmlu}. 
Building upon these advancements, Large Vision-Language Models (LVLMs) \cite{llava,llava1.5} extend LLMs by converting images into hundreds or thousands of sequential visual representations and combined with text prompts. With the advancement of powerful visual understanding and question-answering capabilities, there has also been a rapid increase in computational complexity and memory usage.

A pioneering study FastV \cite{fastv} has highlighted that while image tokens account for the majority of computational load, they receive far less attention than text tokens in the deeper layers of Large Vision-Language Models. A phenomenon known as ``attention anchoring" \cite{streamingllm,fastv} has also been identified, where attention becomes concentrated on just a few specific image tokens, leaving most tokens with substantially reduced focus. Quantitative analyses in some studies \cite{li2024snapkv,fitprune} further demonstrate that core attention points show minimal variation across layers and generation steps, reinforcing this observation. 

Based on the above insights, some methods \cite{fastv, fitprune, avl, xing2024pyramiddrop} have been developed to prune image tokens with low attention scores at specific layers, reducing the computational load to accelerate inference while minimizing the impact on accuracy. 
Image tokens are re-evaluated based on their average received attention scores. Tokens falling below a predefined attention score threshold are then selectively discarded in subsequent layers, streamlining the process by focusing on the most impactful tokens.
They not only bypass the computational demand of the self-attention module but also the Feed-Forward Network (FFN) module in deeper layers.

However, these pruning techniques are typically constrained to around 50\% token reduction, as further pruning often results in substantial performance degradation across tasks. 
As shown in Figure~\ref{fig:figure1}, we demonstrate that attention-only methods represented by FastV \cite{fastv} have a tendency to keep tokens corresponding to the lower part of the image and often select clusters of tokens with high similarity. 
Our in-depth analysis of image tokens has highlighted two primary issues that constrain current attention-based pruning methods: positional bias of image attention and information redundancy in similar image tokens. 
These factors become especially problematic when aiming for more aggressive token pruning, and sharply degrading model performance.

To address these challenges, we propose AdaptPrune, a pioneering training-free pruning method for image tokens that incorporates three critical cues: attention scores, spatial positions, and token similarity. As depicted by the comparison in Figure~\ref{fig:figure1}, in contrast to FastV's tendency to preserve clusters of similar tokens and focus on the lower part of the image, our AdpatPrune selects tokens that better represent diverse semantic regions for more information.

We innovatively reframe the visual token pruning problem as a Non-Maximum Suppression (NMS) task. Drawing inspiration from studies in the object detection area \cite{nms, bodla2017softnms, liu2019adaptivenms, learningnms}, we treat each token as a candidate "detection" point, where its importance score reflects confidence and similarity indicates information density.
AdaptPrune builds on attention scores and develops an adaptive NMS approach, suppressing tokens within a specified range based on spatial distance and token similarity. 
Our method initiates token pruning from a single layer, applying the pruning strategy consistently across subsequent layers and tokens without further adjustments to avoid introducing additional factors that could affect pruning outcomes.

We validate our findings through comprehensive statistical and visual analyses. 
In our experiments, we experiment on 5 well-known LVLMs and evaluate our AdaptPrune on multiple tasks including Image Captioning, General VQA, Text-based VQA, and Multimodal Reasoning.
Extensive experiments demonstrate that AdaptPrune effectively addresses previous limitations, significantly minimizing performance loss even at higher pruning ratios across diverse models and benchmarks.

We summarize our contributions as follows:
\begin{itemize}
    \item We identify and analyze the positional bias and information redundancy phenomena in attention distribution, emphasizing that spatial information and token similarity are also needed for a more holistic strategy.
    \item We propose AdaptPrune, an effective plug-and-play visual token pruning method that reframes the problem as an adaptive NMS task, integrating three critical cues: attention score, spatial distance, and token similarity.
    \item We perform extensive experiments on diverse vision-language benchmarks based on different LVLMs to verify the effectiveness of our AdaptPrune.
\end{itemize}

\section{Related Work}

\textbf{Large Vision-Language Models (LVLMs).} By employing a cross-modal projector that aligns visual encoders \cite{clip} with Large Language Models \cite{llama,devlin2018bert,chiang2023vicuna,qwen,t5,instructblip,sharegpt4v}, LVLMs effectively bridged visual and language understanding \cite{llava,llava1.5,minigpt4}. Despite their effectiveness in various vision-language tasks, LVLMs still encounter challenges such as detailed information understanding and hallucinations \cite{yin2023woodpecker}. To address this, recent advancements focused on increasing image resolution \cite{bai2023qwenvl,chen2024internvl,internvl1_5}, which greatly enhanced the performance on fine-grained tasks like TextVQA \cite{textvqa} and DocVQA \cite{docvqa}. However, these approaches substantially raised the number of image tokens, increasing from 576 in LLaVA-1.5 \cite{llava} to over 2,000 in LLaVA-NEXT \cite{llava1.5} and even exceeding 10,000 in InternVL2 \cite{internvl1_5}, resulting in significantly higher computational demands and memory usage. Even with the KV Cache increasing the speed of the decoding stage, the computational complexity still scales quadratically with the token length in the prefill stage. Consequently, research into the effective pruning of redundant visual tokens becomes essential for optimizing the efficiency of LVLMs in real-world applications.

\textbf{Token Reduction.}
Recent advancements in adaptive attention have improved computational efficiency in LLMs by reducing redundant tokens dynamically \cite{yu2024balancing,votemix,lessismore,fu2024lazyllm,jiang2024minference}. StreamingLLM \cite{streamingllm} retained fixed-position caches for long-context processing, while FastGen \cite{fastgen} adapted KV Cache management to retain important tokens based on recent attention patterns. H2O \cite{zhang2024h2o} and ScissorHands \cite{liu2024scissorhands} further optimized inference by selectively pruning KV pairs and retaining tokens with consistently high attention. 
Many trainable methods \cite{kong2022spvit, llavaprumerge, chen2024llavolta, liang2022not, xiong2025pyra, ding2023prune, shi2023crossget} focused on visual token reduction in vision transformers \cite{clip} before image tokens are projected into LLMs. However, when directly applied to LVLMs without training, these methods lead to performance degradation because they fail to incorporate LLM's understanding of the textual context.

FastV \cite{fastv} proposed an image token pruning method for LVLMs to address inefficiencies in image attention. This method pruned tokens based on their attention scores starting from a specific layer and achieved minimal impact on overall performance. FastV \cite{fastv} has been validated across various benchmarks and models, with additional solutions introduced ensuring compatibility with KV Cache. Later methods \cite{vtw,arif2024hired,lookm,fitprune,avl,xing2024pyramiddrop,elasticcache} explored dynamic pruning for each token or multi-layer pruning based on attention scores. Despite its widespread usage in existing methods, the ranking criteria for token pruning have received limited exploration, which we aim to further improve.

Unlike most methods developed after FastV, which focus on multilayer or dynamic pruning for different output tokens, we aim to explore a more fundamental and crucial aspect of token pruning which is the cues of token ranking. By addressing the phenomena we identified, We develop a more holistic ranking standard for visual token pruning. The detailed elaboration follows below.

\section{Observations and Insights}
In this section, we present several phenomena observed through experiments across different datasets. Prior work such as FastV \cite{fastv} has revealed that image tokens constitute a substantial portion of input tokens, and they receive disproportionately less attention in deeper layers. 
Further studies \cite{fastv, li2024snapkv, avl, zhang2024pyramidkv} have shown that the model consistently focuses on a subset of image tokens across multiple layers and generation steps.
Building upon these significant findings, our experimental observations lead to new insights into the attention mechanisms of LVLMs, motivating us to propose strategies to improve token selection.

\begin{figure}[t]
  \vspace{-10pt}
  \centering
  \includegraphics[width=\linewidth]{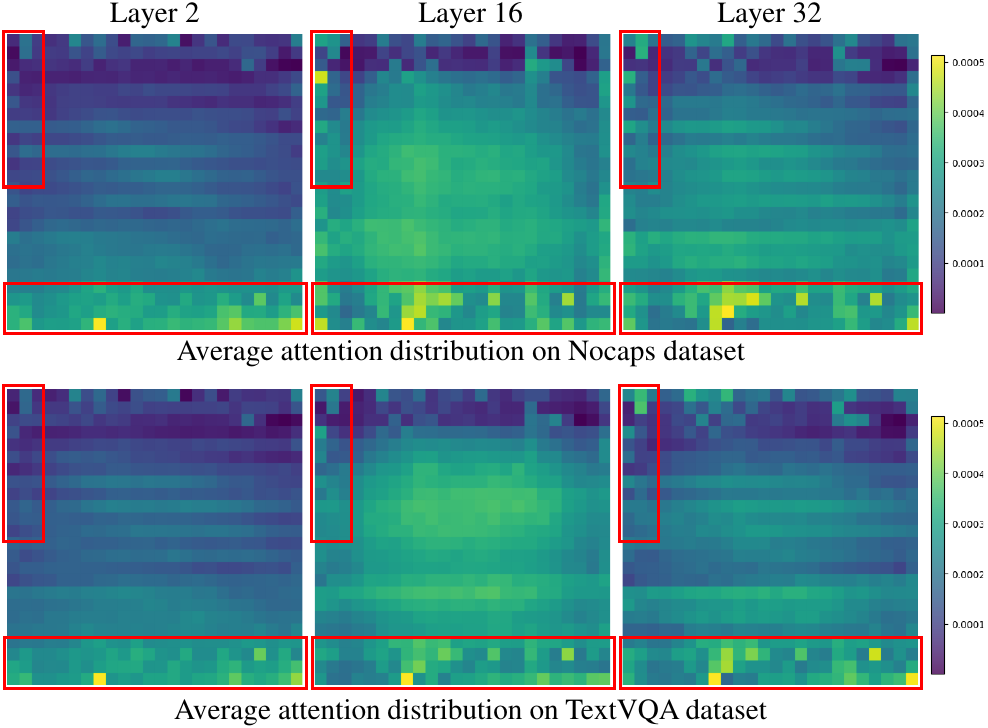}
  \caption{
        Visualization of attention positional bias phenomenon. The average attention distribution projected onto image patches exhibits a fixed pattern across different layers and tasks.
  }
  \label{fig:observe3}
  \vspace{-0.1in}
\end{figure}

\textbf{Positional Bias in Attention Distribution.} 
As shown in Figure~\ref{fig:observe3}, we observe fixed motifs and patterns in the attention score distribution of image tokens which will cause positional bias in token pruning. We use 1000 samples from the Nocaps \cite{agrawal2019nocaps} dataset of the Image Captioning task and the TextVQA \cite{textvqa} dataset of the Text-based VQA task for statistics. 
Despite these samples coming from entirely different tasks, the model exhibits an almost identical attention distribution motifs for image tokens on the same layer.
Across different layers, the shapes changes slightly but the tokens with the higher attention remain the same. 

We observe that higher attention is distributed at the end of an image sequence. We interpret this phenomenon as a result of LVLMs inheriting the attention characteristics of their LLM component. Image patches are encoded as a token sequence and processed almost the same as a text paragraph during the pipeline of LVLMs. However, unlike text tokens, where boundary locations usually contain key information, the key information of an image is usually located in the central region \cite{gaussian}. 
Pruning methods that rely solely on attention scores \cite{fastv,avl,fitprune,xing2024pyramiddrop} tend to keep tokens in certain fixed positions and discard other essential tokens due to the bias of the model. As a result, such pruning strategies lead to performance degradation as the pruning ratio increases.

\begin{figure}[t]
  \vspace{-10pt} 
  \centering
  \includegraphics[width=\linewidth]{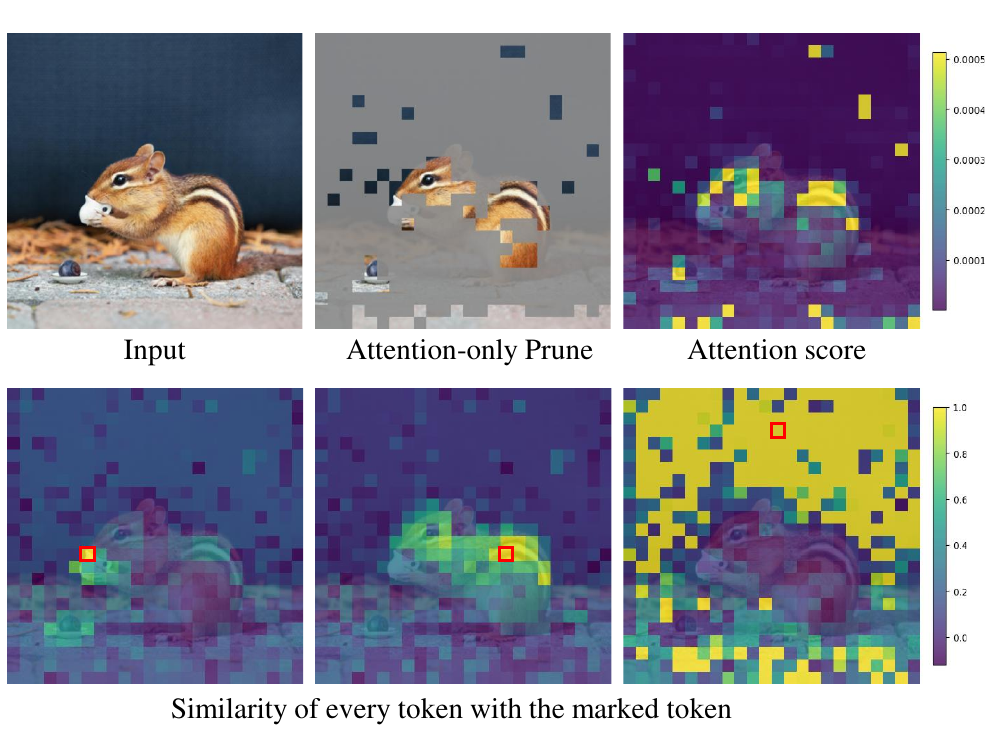}
  \caption{
        Visualization of the information redundancy phenomenon. The tokens with the highest attention score are kept, most of which are adjacent and have high similarity.
  }
  \label{fig:observe3_4}
  \vspace{-0.1in}
\end{figure}

\begin{figure*}[t]
  \centering
  \includegraphics[width=\linewidth]{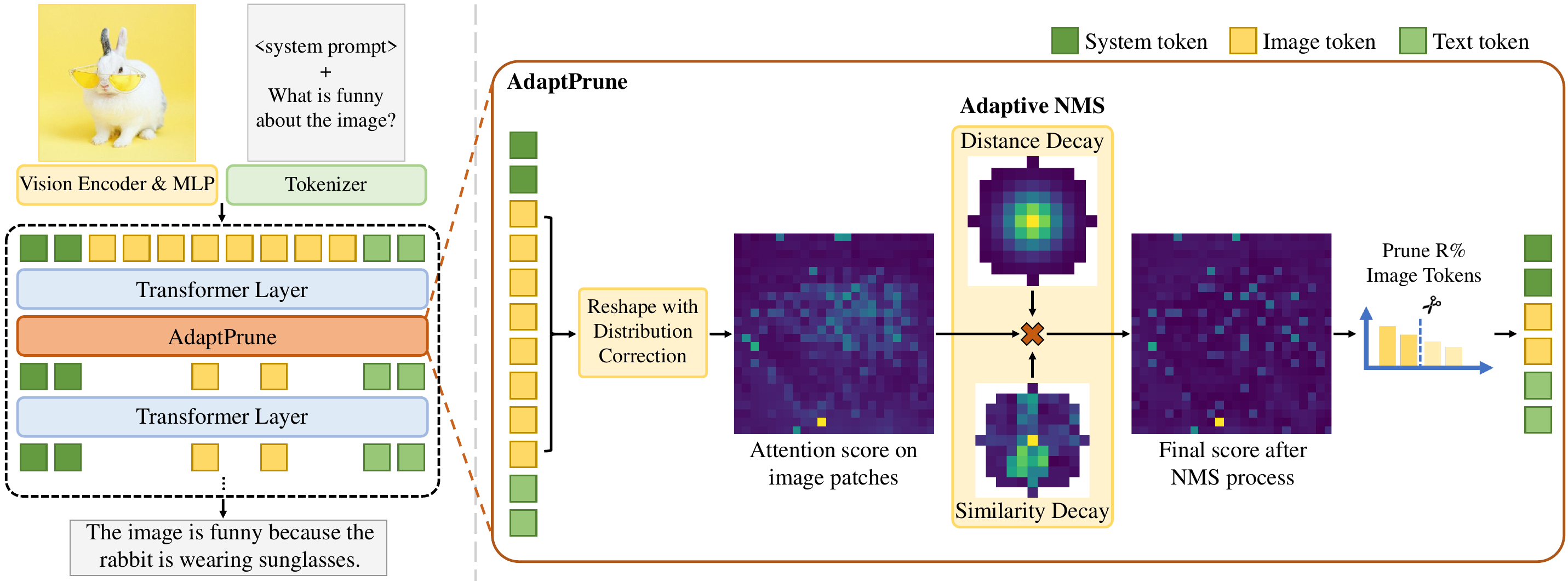}
  
  \caption{An overview of our AdaptPrune framework. AdaptPrune employs a simple and effective one-layer pruning strategy and incorporates an adaptive NMS algorithm to optimize the pruning process. AdaptPrune dynamically adjusts the suppression process using three critical cues: attention score, spatial distance, and token similarity.
  }
  \label{fig:framework}
\end{figure*}

\textbf{Information Redundancy in High Attention Tokens.} 
When token selection relies solely on attention scores, the model tends to retain clusters of similar tokens, resulting in redundant information and inefficient use of computational resources. As shown in Figure~\ref{fig:observe3_4}, adjacent tokens with similar visual features in LVLMs often receive similar attention scores, particularly in areas with flat backgrounds or repetitive textures. Spatially close tokens capture overlapping features, making it difficult for the attention mechanism to differentiate unique or critical information. While attention-based pruning effectively captures central focus areas, it often overlooks important background context, leading to wasted attention on redundant information. 
To improve pruning efficiency and model performance, it is essential to incorporate similarity and informational uniqueness into the token selection process.

\section{Method}

As shown in Figure~\ref{fig:framework}, our AdaptPrune reframes visual token pruning as a Non-Maximum Suppression (NMS) problem. Drawing inspiration from object detection \cite{nms,bodla2017softnms,liu2019adaptivenms}, we treat each token as a candidate “detection” point with an importance score, which serves as a confidence measure. In our AdaptPrune, tokens with high similarity and proximity to the selected token are suppressed, retaining only diverse and essential information for downstream tasks. AdaptPrune optimizes inference by preserving token diversity and reducing redundancy, ultimately enhancing computational efficiency without sacrificing model performance.

\subsection{Distribution Correction}

Based on our findings of attention's positional bias, we propose adjusting the distribution of attention scores across image patches. Figure 8 of an influential study \cite{gaussian} points out that more information is distributed around the center of the image and suggests that the Gaussian function provides a good approximation. We apply a Gaussian mask centered at the image's midpoint to the attention scores. This correction results in a more balanced attention distribution.

\subsection{Factor Measurement}
We employ three essential cues to calculate the scores of tokens in different aspects, explained below in terms of their measurement and computation.

\textbf{Attention Scores Measurement.}
Like many existing methods \cite{fastv}, we utilize the attention values from the previous layer in the LVLM's current token generation step to represent the importance of image tokens. 
This approach intuitively reflects the dependency of the current output token on all visual tokens and has been demonstrated by numerous prior studies to be an effective scoring mechanism. 

\textbf{Distance Measurement.}
Our method refines attention scoring by applying Soft NMS with two-dimensional spatial distance. To calculate this distance, we map image tokens back to their original patch positions. Unlike traditional NMS, which directly removes overlapping areas, Soft NMS gradually reduces scores based on token proximity, which prevents the forced removal of dense neighboring information. By leveraging a Gaussian decay function, we can control the range and intensity of distance suppression. This approach more efficiently chooses diverse regions across the image to capture representative information.

\textbf{Token Similarity Measurement.} 
In high-resolution Text-based VQA benchmarks such as TextVQA \cite{textvqa} and DocVQA \cite{docvqa}, answers are often concentrated in small, high-density regions, with each token representing unique details of the answer. This contrasts with tasks like Image Captioning or General VQA, which require broader contextual understanding. Drawing inspiration from the Zero-TPrune \cite{zerotprune} method, we incorporate token similarity as an additional filtering factor. By calculating the cosine similarity between key vectors, we implement an adaptive NMS that dynamically adjusts decay parameters based on both spatial distance and similarity. This adaptive pruning strategy effectively removes redundant tokens while retaining critical ones, enabling the model to better capture essential details at any density within the answer region.

\subsection{Adaptive NMS}
We propose an adaptive NMS algorithm suitable for candidate token selection in visual token pruning to integrate these three essential cues.
At each iteration of the NMS process, we select the token \( t_i \) with the highest attention score \( s_i \). For the currently selected token, we apply a suppression function to adjust the scores of nearby tokens \( t_j \) within a spatial range. The adaptive suppression function combines decay factors based on spatial distance and token similarity.

We compute the spatial decay function \( D_{\text{spatial}} \) and similarity decay function \( D_{\text{similarity}} \) as follows:
\begin{equation}
D_{\text{spatial}} = \exp \left(-\frac{d(t_i, t_j)^2}{2\sigma_d^2}\right),
\end{equation}
\begin{equation}
D_{\text{similarity}} = \exp \left(-\frac{(1 - \cos(\text{key}_i, \text{key}_j))^2}{2\sigma_s^2}\right).
\end{equation}
The decay functions are calculated using spatial distance \( d(t_i, t_j) \) and cosine similarity \( \cos(\text{key}_i, \text{key}_j) \) between key vectors of the tokens, each modulated by decay factors \( \sigma_d \) and \( \sigma_s \), respectively. The updated score \( s_j \) for each nearby token \( t_j \) is then calculated as:
\begin{equation}
s_j \mathrel{\mathop{:}{=}}
 s_j \cdot \left(1 - D_{\text{spatial}} \cdot D_{\text{similarity}}\right).
\end{equation}

This function adaptively suppresses token scores using a multiplicative decay. Initial scores \( s_j \) are adjusted so that \( s_j \) reflects its spatial proximity and feature resemblance to \( t_i \). This approach prioritizes spatially sparse tokens with distinct features, enhancing token selection flexibility through the adaptable decay parameters \( \sigma_d \) and \( \sigma_s \).

At each step, we select the highest-scoring token that hasn’t yet been chosen and retain it. We then suppress the scores of surrounding tokens based on their spatial distance and similarity to the selected token. This iterative process continues until the pruning preservation ratio is met, with each newly selected token reducing the scores of its neighbors. This approach ensures that we retain only the most influential and representative tokens while effectively pruning those with redundant or less impactful information. Algorithm 1 shows the pseudocode for AdaptPrune.

\subsection{Discussion}

To summarize, our AdaptPrune introduces an advanced training-free token pruning strategy that unifies the cues of attention scores, spatial distance, and token similarity into cohesive importance ranking criteria of a balanced pruning approach. Applying AdaptPrune during the prefilling phase accelerates the prefilling speed while reducing the KV Cache length, which further enhances the decoding process. Our computational efficiency is almost identical to that of FastV \cite{fastv}, as both are single-layer pruning methods. 
Our adaptive strategy minimizes floating-point operations per second (FLOPs), effectively enhancing both inference speed and memory efficiency.

\begin{algorithm}[t]
\caption{AdaptPrune}
\begin{algorithmic}[1]
\Require Tokens $T$, scores $S$, positions $P$, keys $K$, distance decay parameter $\sigma_d$, similarity decay parameter $\sigma_s$, pruning ratio $\text{ratio}$.
\Ensure Retained tokens $R$.
\State $R \gets \emptyset$;
\State $N \gets \text{ratio} \times |T|$;
\State $S \gets \text{GaussianCorrection}(S, P)$;
\While{$|R| < N$}
    \State $i \gets \arg \max_{k \in T} S[k]$;
    \State $R \gets R \cup \{T[i]\}$;
    \For{$j \in \text{Neighbors}(i)$}
        \State $d \gets \|P[i] - P[j]\|$;
        \State $sim \gets \cos(K[i], K[j])$;
        \State $D_{\text{spatial}} \gets \exp\left(- \frac{d^2}{2\sigma_d^2}\right)$;
        \State $D_{\text{similarity}} \gets \exp\left(- \frac{(1 - sim)^2}{2\sigma_s^2}\right)$;
        \State $S[j] \gets S[j] \cdot \left(1 - D_{\text{spatial}} \cdot D_{\text{similarity}}\right)$;
    \EndFor
\EndWhile
\State \Return $R$.
\end{algorithmic}
\end{algorithm}

Unlike previous methods that primarily consider attention scores, AdaptPrune leverages spatial and similarity information to retain a more diverse token set, ensuring minimal impact on model accuracy even with substantial token reduction. By benefiting both the prefilling and decoding phases, this method offers a comprehensive optimization for long-context multimodal tasks, balancing memory efficiency and high-quality output retention.

\begin{table*}[t]
\setlength{\tabcolsep}{0.85mm}
\centering
\small
  
  \begin{tabular}{l|cc|cccc|ccc|ccc}
    \toprule
    \multirow{2}{*}{Method} & \multicolumn{2}{c|}{Image Caption} &  \multicolumn{4}{c|}{General VQA} & \multicolumn{3}{c|}{Text-based VQA} &   \multicolumn{3}{c}{ Multimodal Reasoning}  \\
    & Nocaps & Flickr30K & VQAv2 & GQA  & OK-VQA & VizWiz & TextVQA & ChartQA & DocVQA & POPE & MME & MMB\\ 

\midrule
LLaVA-1.5-7B \cite{llava1.5}  & 105.8  & 75.2 & 76.6  & 61.9 & 53.4 & 54.2 & 45.9 & 18.1 & 28.1 & 87.0 & 1508.2 & 64.1 \\
\ + VTW \cite{vtw} (K=5)           & 6.1 & 6.9 & 42.7 & 39.5 & 20.0 & 50.3 & 8.1 & 11.6 & 10.5 & 46.0 & 635.5 & 21.0  \\
 \ + FastV \cite{fastv}            & 74.2   & 47.4 & 63.8  & 51.0 & 43.9 & 54.9 & 38.5 & 14.6 & 18.0 & 69.7 & 1217.8 & 58.6 \\ 
               \ + \textbf{AdaptPrune}      & \textbf{99.3}   & \textbf{70.5} & \textbf{72.3}  & \textbf{58.0} & \textbf{49.1} & \textbf{55.9} & \textbf{40.1} & \textbf{17.0} & \textbf{19.4} & \textbf{79.3} & \textbf{1393.2} & \textbf{59.4} \\ 
\midrule
LLaVA-1.5-13B \cite{llava1.5} & 109.4  & 79.5 & 79.2  & 63.3 & 58.3 & 56.6 & 48.7 & 18.1 & 30.3 & 87.1 & 1528.8 & 68.7 \\ 
\ + VTW \cite{vtw} (K=5)           & 5.0 & 4.8 & 47.0 & 39.7 & 22.4 & 50.2 & 8.2 & 13.2 & 11.0 & 50.0 & 623.7 & 22.4  \\
                 \ + FastV \cite{fastv}         & 94.6   & 65.4 & 68.7  & 54.6 & 51.4 & 57.2 & 38.3 & 15.2 & 17.5 & 74.8 & 1353.4 & 63.3 \\ 
                 \ + \textbf{AdaptPrune}    & \textbf{103.8}  & \textbf{75.2} & \textbf{73.5}  & \textbf{57.9} & \textbf{53.1} & \textbf{57.9} & \textbf{42.0} & \textbf{15.4} & \textbf{19.8} & \textbf{80.5} & \textbf{1461.2} & \textbf{65.1} \\ 
\midrule
LLaVA-NEXT-7B \cite{llava1.5} & 88.3   & 68.4 & 81.8  & 64.2 & 44.2 & 60.6 & 64.8 & 54.8 & 74.4 & 87.6 & 1519.5 & 67.2 \\ 
                 \ + FastV \cite{fastv}        & 72.7   & 52.2 & 71.3  & 56.6 & 38.3 & 57.4 & 48.2 & 30.0 & 36.8 & 79.7 & 1352.7 & 61.3 \\ 
                \ + \textbf{AdaptPrune}     & \textbf{82.1}   & \textbf{61.7} & \textbf{76.7}  & \textbf{59.6} & \textbf{41.0} & \textbf{58.9} & \textbf{55.8} & \textbf{33.8} & \textbf{45.0} & \textbf{84.3} & \textbf{1428.3} & \textbf{63.0} \\ 

    \bottomrule
  \end{tabular}
  \caption{
  Experiments on LLaVA with 90\% pruning ratio for visual tokens. Under this experiment setting, the FLOPs of the LLaVA-1.5 \cite{llava1.5} baseline model are reduced by around 81\%, while the FLOPs of the LLaVA-NEXT \cite{llava1.5} baseline model are reduced by around 87\%.
  }
  \label{tab:llavatable}
\end{table*}

\begin{table*}[t]
\setlength{\tabcolsep}{2.2mm}
\centering
\small
  
  \begin{tabular}{l|cccc|ccc|ccc}
    \toprule
    \multirow{2}{*}{Method} &  \multicolumn{4}{c|}{General VQA} & \multicolumn{3}{c|}{Text-based VQA} &   \multicolumn{3}{c}{ Multimodal Reasoning}  \\
    & VQAv2 & GQA & OK-VQA & VizWiz & TextVQA & ChartQA & DocVQA & POPE & MME & MMB \\
\midrule
InternVL2-2B \cite{internvl1_5} & 74.7 & 59.9 & 43.6 & 46.1 & 72.7 & 74.9 & 87.4 & 89.0 & 1444.6 & 72.3 \\ 
\ + FastV   \cite{fastv}         & 64.3 & 50.9 & 37.1 & 42.8 & 49.9 & 17.9 & 28.0 & 80.1 & \textbf{1339.3} & 65.2 \\ 
\ + \textbf{AdaptPrune}       & \textbf{66.5} & \textbf{52.1} & \textbf{39.2} & \textbf{44.2} & \textbf{52.7} & \textbf{22.8} & \textbf{31.1} & \textbf{82.5} & 1313.5 & \textbf{68.2} \\ 
\midrule
InternVL2-8B \cite{internvl1_5} & 78.9 & 62.7 & 52.2 & 61.0 & 77.0 & 82.4 & 92.0 & 87.9 & 1628.7 & 82.0 \\ 
\ + FastV  \cite{fastv}          & 67.4 & 53.6 & 45.3 & 58.2 & 55.7 & 28.2 & 33.8 & 79.6 & 1510.0 & 71.2 \\ 
~ + \textbf{AdaptPrune}       & \textbf{71.6} & \textbf{55.7} & \textbf{47.5} & \textbf{59.9} & \textbf{57.4} & \textbf{32.8} & \textbf{42.9} & \textbf{82.1} & \textbf{1516.7} & \textbf{75.6} \\ 
    \bottomrule
  \end{tabular}
  \caption{
  Experiments on InternVL with 90\% pruning ratio for visual tokens. InternVL \cite{internvl1_5} model FLOPs are reduced by around 88\%.
  }
  \label{tab:internvltable}
  \vspace{-0.1in}
\end{table*}

\section{Experiments}

We evaluate the effectiveness of our method across 5 different models and 12 different benchmarks. We aim to provide a comprehensive comparison of the performance of visual token pruning across diverse multimodal tasks.

\begin{figure*}[t]
  \centering
  \includegraphics[width=\linewidth]{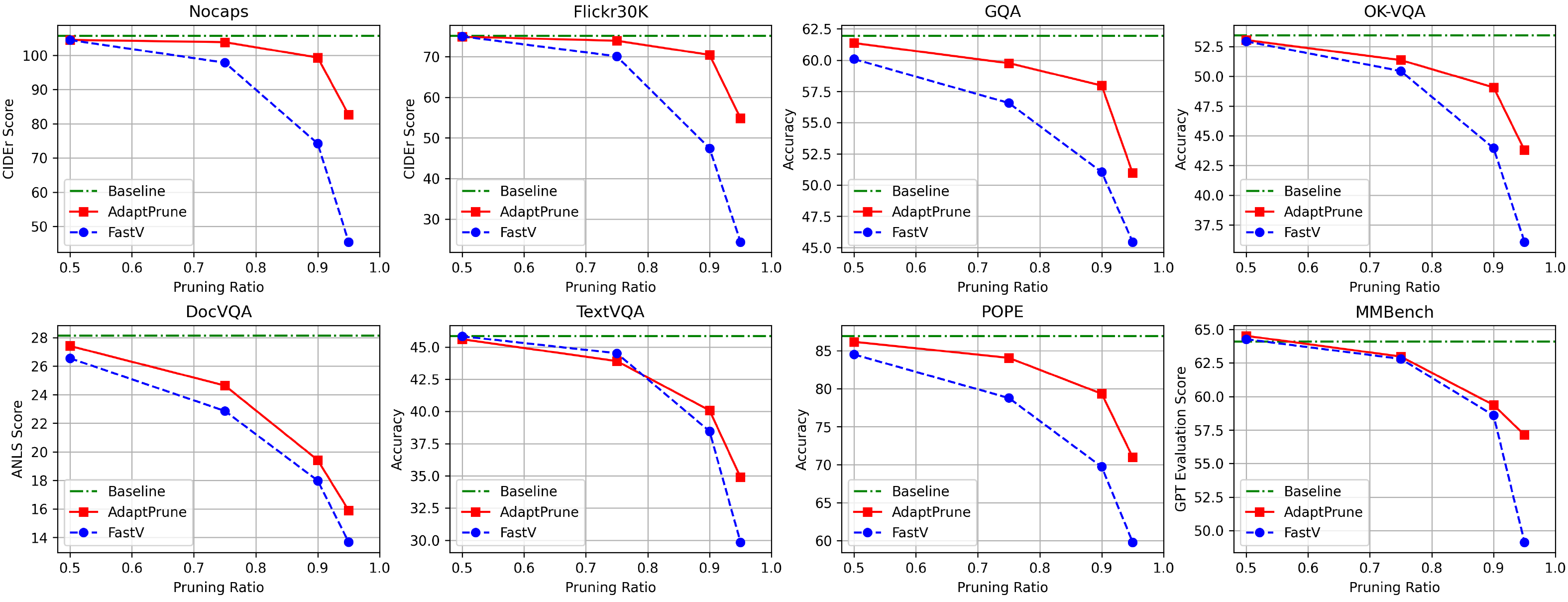}
  
  \caption{
  Performance comparison of  FastV \cite{fastv} and AdaptPrune on the LLaVA-1.5-7B model at different pruning ratios across eight evaluation benchmarks. 
  The green line represents the LLaVA-1.5-7B baseline results without pruning.
  }
  \vspace{-0.09in}
  \label{fig:ratio}
\end{figure*}

\begin{figure}[t]
  \centering
  \includegraphics[width=\linewidth]{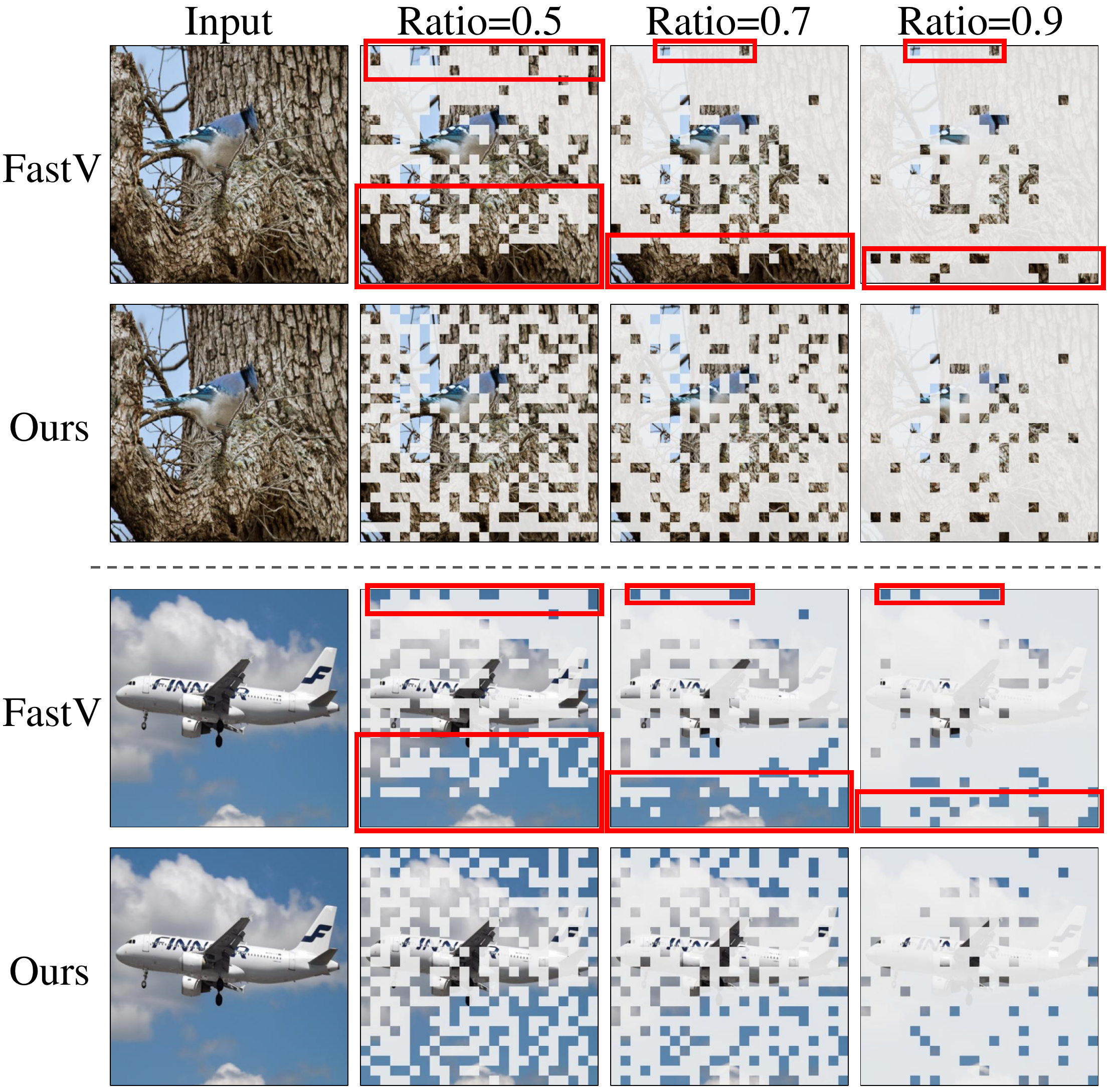}
  \caption{
        Visualization comparison between FastV [9] and our AdaptPrune on LLaVA-1.5-7B \cite{llava1.5} at different pruning ratios of samples from Nocaps (upper) and TextVQA (lower). FastV's redundant tokens at fixed positions are marked with red bboxes.
  }
  \vspace{-0.15in}
  \label{fig:ratio_nocaps}
\end{figure}

\subsection{Model Settings}

For a fair comparison and efficient validation, we employ the LMMs-Eval \cite{zhang2024lmmseval} evaluation framework across all experiments. The models used include LLaVA-1.5 (7B and 13B variants) \cite{llava1.5}, LLaVA-NEXT-7B \cite{internvl1_5}, and InternVL2 (2B and 8B variants) \cite{internvl1_5}, with configurations based on the their official settings. Detailed model architectures and settings are provided in the supplementary material.

\subsection{Datasets}

To demonstrate the broad applicability of our approach, we use several representative benchmark datasets that cover a wide spectrum of task domains for evaluation, including Nocaps \cite{agrawal2019nocaps} and Flickr30K \cite{plummer2015flickr30k} for Image Captioning, VQAv2 \cite{vqav2}, GQA \cite{hudson2019gqa}, OK-VQA \cite{marino2019okvqa} and VizWiz \cite{gurari2018vizwiz} for General Visual Question Answering, TextVQA \cite{textvqa}, ChartQA \cite{masry2022chartqa} and DocVQA \cite{docvqa} for Text-based Visual Question Answering and POPE \cite{pope}, MME and MMBench-EN \cite{liu2025mmbench} for Multimodal Reasoning. Detailed descriptions of the datasets and corresponding evaluation metrics are provided in the supplementary material.

\subsection{Main Results}
We begin by comparing the pruning result of FastV \cite{fastv} and AdaptPrune across 5 LVLMs and 12 benchmarks at a high pruning ratio. Then we show experiments with various pruning ratios and qualitative visualizations to showcase the pruning effects of our method. To ensure a fair comparison, we align key variables such as the pruning layer and pruning ratio across both methods. Specifically, the pruning layer is set to 3 following the optimal setup of FastV \cite{fastv}.

\textbf{Comparison on 90\% Pruning Ratio.} As shown in Tables \ref{tab:llavatable} \ref{tab:internvltable} and \ref{tab:speed}, we evaluate 5 LVLMs using VTW \cite{vtw}, FastV \cite{fastv} and AdaptPrune. Since AdaptPrune introduces a new ranking criterion in token selection process, we compare its single-layer pruning performance with FastV and VTW \cite{vtw} for a fair evaluation. The evaluated LVLMs differ in model size, training data, and architecture. Since the InternVL2 \cite{internvl1_5} framework does not include Image Captioning tasks with CIDEr score \cite{vedantam2015cider}, we omit those results. To assess method resilience, we select a 90\% pruning ratio to test token pruning effectiveness, emphasizing AdaptPrune’s ability to maintain performance with minimal token retention, unlike FastV, which struggles with aggressive pruning.

Our method outperforms FastV across all models and benchmarks. Unlike previous attention-only methods that suffer performance drops at high pruning ratios, our approach minimizes loss even under aggressive pruning.

Experiments on Nocaps \cite{agrawal2019nocaps}, Flickr30k \cite{plummer2015flickr30k}, and General VQA benchmarks show that our method retains more informative tokens, preserving broader image content. As shown in Figure \ref{fig:ratio} and \ref{fig:ratio_nocaps}, while FastV \cite{fastv} suffers significant losses in image semantic information under extreme pruning, our analysis suggests that the semantic loss primarily stems from suboptimal ranking functions. Many non-informative tokens are included in the retained token set and genuinely informative tokens are filtered out, which causes the main performance decline. Our new pruning strategy selects a more informative and representative token set and preserves a wider array of image information.

For Text-based VQA benchmarks like TextVQA \cite{textvqa}, DocVQA \cite{docvqa}, and ChartQA \cite{masry2022chartqa}, where answers are often confined to small, information-dense regions, the pruning algorithm must measure token density and redundancy. This necessitates that the pruning algorithm should effectively measure both the density and redundancy of token information. By incorporating token similarity into our AdaptPrune, we dynamically control the suppression degree, adjusting to varying levels of information clustering and redundancy. This demonstrates the effectiveness of the AdaptPrune strategy in leveraging multiple pruning cues to preserve model performance across various conditions.

\textbf{Comparison at Different Pruning Ratios.} As shown in Figure~\ref{fig:ratio} We conduct experiments across a range of token pruning ratios, from 0.5 to 0.95, to evaluate how well our method maintains model performance at varying levels of token retention. By testing across these ratios, we aim to understand not only the robustness of our approach under different pruning intensities but also its adaptability across models and benchmarks. Unlike prior approaches that show steep performance degradation as pruning becomes more aggressive, our method demonstrates flexibility by preserving key image semantic information, even at high pruning ratios. This suggests that our method can dynamically adjust to different levels of token sparsity, maintaining the quality of outputs while improving efficiency.

\begin{table}[t]
\setlength{\tabcolsep}{1.2mm}
\centering
\small
  
  \vspace{-0.02in}
  \begin{tabular}{l|cccc}
    \toprule
     Method & Time & GPU-Memory  & Score & Latency\\
     
    \midrule
    LLaVA-1.5-7B \cite{llava1.5} & 49:05 & 19G  & 87.0 & 0.327s \\
    \ + FastV \cite{fastv} &  27:30 & 14.4G & 69.7 & 0.182s \\
    \ + \textbf{AdaptPrune} & 31:25 & 14.5G  & 79.3 & 0.209s \\
    
    \bottomrule
  \end{tabular}
  \caption{Real inference comparison on POPE dataset where the output is only one token. Experiments are on LLaVA-1.5-7B \cite{llava1.5} on single 3090 GPU with the pruning ratio of 90\%. 
  }
  \label{tab:speed}
  \vspace{-0.15in}
\end{table}

\textbf{Qualitative Comparison.}
As shown in Figure \ref{fig:ratio_nocaps}, to gain deeper insights into the behavior of our method, we conduct a detailed visualization analysis using the LLaVA-1.5-7B model across all benchmarks.  By sampling 1,000 images, we can examine token selection and pruning patterns in a more granular way.  This analysis reveals how our approach differs from previous methods in selecting informative tokens and filtering out redundant ones.  Specifically, we observe that our method consistently retains tokens that capture critical spatial and semantic details, leading to richer and more contextually accurate outputs.  These visual comparisons illustrate how our adaptive token pruning strategy enhances model efficiency without compromising on the essential information needed for high-quality predictions.

\begin{table*}[t]
\setlength{\tabcolsep}{1.2mm}
\centering
\small

  \begin{tabular}{c|ccc|cccccccc}
    \toprule
    
    Method &  Correction & Distance & Similarity & Nocaps & Flickr30K  & GQA & OK-VQA & DocVQA & TextVQA & POPE & MMB\\
     
    \midrule
    LLaVA-1.5-7B \cite{llava1.5} & & & & 105.8 & 75.2 &  61.9 & 53.4  & 28.1 & 45.9 & 87.0 & 64.1 \\
    \midrule
    (a) & & & & 74.2 & 47.5 & 51.0 & 43.9 & 18.0 & 38.5 & 69.7 & 58.6 \\
    (b) & \checkmark &  & &   95.9 & 68.8 & 55.4 & 48.2 & 18.2 & 37.5 & 77.9 & 59.8 \\
    (c) & \checkmark & \checkmark & &  98.9 & 70.0 & 56.2 & 47.1  & 17.3 & 36.9 & 78.8 & \textbf{60.1} \\
    (d) & \checkmark & \checkmark & \checkmark & \textbf{99.3} & \textbf{70.5} & \textbf{58.0} & \textbf{49.1}  & \textbf{19.4} & \textbf{40.1} & \textbf{79.3} & 59.4 \\
    
    \bottomrule
  \end{tabular}
  \caption{Ablation study across factors introduced in our AdaptPrune. Method (a) represents FastV \cite{fastv} and the term ``Correction" stands for our Distribution Correction operation. Experiments are conducted on LLaVA-1.5-7B \cite{llava1.5} with the pruning ratio of 90\%.
  }
  \label{tab:ablation_1}
  \vspace{-0.07in}
\end{table*}

\begin{table}[t]
\setlength{\tabcolsep}{1.2mm}
\centering
\small
  
  \vspace{-0.02in}
  \begin{tabular}{l|cccc}
    \toprule
     Method & GQA & OK-VQA  & TextVQA & MMB\\
     
    \midrule
    LLaVA-1.5-7B \cite{llava1.5} & 61.9 & 53.4  & 45.9 & 64.1 \\
    \ + FastV \cite{fastv} &  51.0 & 43.9 & 38.5 & 58.6 \\
    \ + \textbf{AdaptPrune} & \textbf{58.0} & \textbf{49.1}  & \textbf{40.1} & \textbf{59.4} \\
    \midrule
    (a) + FitPrune \cite{fitprune}  & 45.9 & 32.3 & 27.0 & 47.3\\
    (b) + Skip  & 54.7 & 45.4  & 35.3  & 59.1  \\
    (c) + Random  & 56.0 & 40.0 & 20.2 & 56.7\\
    (d) + Random 3x3  & 56.2 & 42.2 & 22.7 & 56.2\\
    (e) + Max-pooling 3x3 & 56.0 & 47.7 & 33.2 & 58.9  \\
    (f) + Average-pooling  & 48.4 & 38.1 & 32.1 & 54.7 \\
    (g) + The last 10\% & 41.3 & 18.6 & 9.2 & 27.5 \\
    
    \bottomrule
  \end{tabular}
  \caption{Ablation study on alternative pruning strategies. Experiments are on LLaVA-1.5-7B \cite{llava1.5} with the pruning ratio of 90\%.
  }
  \label{tab:ablation_2}
  \vspace{-0.1in}
\end{table}

\subsection{Ablation Studies}

To validate the effectiveness across the factors we introduced in our AdaptPrune, we conduct in-depth ablation experiments as detailed in Table \ref{tab:ablation_1} and \ref{tab:ablation_2}. All ablation experiments are conducted on LLaVA-1.5-7B and the performance is assessed across benchmarks from different tasks. 

\textbf{Impact of Distribution Correction.} We first evaluate a simplified variant (Table~\ref{tab:ablation_1} (b)) of our AdaptPrune, where we only apply a Distribution Correction process to minimize the positional bias effect we observed. 
Following this correction, we prune tokens based on the adjusted attention scores. Compared with FastV \cite{fastv} that used the attention score directly (Table~\ref{tab:ablation_1} (a)), this approach improves performance across all benchmarks. This validates our observation that the model's positional attention bias phenomenon. The biased attention distribution does not accurately reflect the true importance of information. 
The performance improvements across all benchmarks demonstrate that the Distribution Correction process allows new scores to more accurately represent the importance of visual information.

\textbf{Impact of Distance Decay.} Based on the simplified variant (Table~\ref{tab:ablation_1} (b)), we further introduce the distance decay to reduce the redundancy (Table~\ref{tab:ablation_1} (c)). We observed that the selected tokens tend to cluster together, which usually means the information is highly similar. To address this, we incorporate the distance factor into our attention-based pruning strategy using a Soft NMS approach. This method suppresses the scores of tokens in the surrounding area of each selected token based on their distance, effectively promoting a more even token distribution across the image.
Experiment results show significant performance improvements in Image Captioning, General VQA, and Multimodal Reasoning tasks. However, on Text-based VQA benchmarks TextVQA \cite{textvqa} and DocVQA \cite{docvqa}, performance is below the attention-only approach of FastV \cite{fastv}. We attribute this to the enforced spatial dispersion of selected tokens, which leads to information loss in tasks where answers are concentrated in very small regions.

\textbf{Impact of Similarity Decay.}
To accommodate tasks with varying information densities, we integrate similarity information into our method (Table~\ref{tab:ablation_1} (d)). To ensure each retained token represents distinct information, we adjust the decay effect on neighboring tokens based on the similarity of their key vectors. Thus we propose the full version of our AdaptPrune approach. AdaptPrune suppresses the surrounding tokens' attention score of the selected tokens with distance and similarity information. Experimental results show that our AdaptPrune consistently outperforms FastV \cite{fastv} across all tasks and achieves the best performance compared to incomplete versions on almost all benchmarks.

\textbf{Comparison with Alternative Pruning Strategy.}
As shown in Table \ref{tab:ablation_2}, we experimented with alternative scoring and pruning strategies under our framework to validate the effectiveness of our method.
We performed a comprehensive comparison with methods such as average pooling, max pooling, and random selection, with the specific implementations detailed in the supplementary material. 
The experimental results demonstrate that our method achieves the highest performance at the same pruning ratio.

\section{Conclusion}

In this paper, we introduce AdaptPrune, a novel training-free token pruning method designed to enhance the efficiency of Large Vision-Language Models (LVLMs). Our method addresses the key limitations of existing token pruning strategies, which predominantly rely on attention scores alone. By incorporating spatial information and token similarity, AdaptPrune provides a more holistic approach to token selection, ensuring that redundant tokens are suppressed while retaining those that contribute to the model’s performance. Through extensive experiments across multiple models and benchmarks, we demonstrate that AdaptPrune significantly improves both speed and memory efficiency while maintaining minimal accuracy loss, even when pruning up to 90\% of the tokens. Our results validate the importance of considering multiple cues of token relevance in optimizing the performance of multimodal models.

{
    \small
    \bibliographystyle{ieeenat_fullname}
    \bibliography{main}
}

\maketitlesupplementary
\appendix

\setcounter{page}{1}

\newpage

In the supplementary material, we provide detailed descriptions of the model architectures and settings, benchmark datasets along with their respective evaluation metrics, and alternative pruning strategies. Additionally, we present fine-grained results on the MMBench and MME benchmarks. We also offer visualization comparisons between FastV \cite{fastv} and our AdaptPrune method at various pruning ratios, along with the average attention distribution across each layer of LLaVA-1.5-7B \cite{llava1.5} on different datasets.

\section{Model Settings}
For a fair comparison and efficient validation, we employ the LMMs-Eval \cite{zhang2024lmmseval} evaluation framework across all experiments. The models used include LLaVA-1.5 (7B and 13B variants) \cite{llava1.5}, LLaVA-NEXT-7B \cite{internvl1_5}, and InternVL2 (2B and 8B variants) \cite{internvl1_5}. We set the configurations based on their official settings.

\textbf{LLaVA-1.5.} The model architecture of LLaVA-1.5 \cite{llava1.5} replaces the linear projection in LLaVA \cite{llava} with an MLP to map visual features into a shared embedding space with the language model. It uses CLIP-ViT-L as the visual encoder at a resolution of 336×336, with Vicuna as the language decoder. The image is divided into 24×24 patches, generating 576 image tokens. We implement our AdaptPrune by restoring the image sequences to a two-dimensional arrangement on LLaVA-1.5.

\textbf{LLaVA-NEXT.} LLaVA-NEXT \cite{llava1.5} is an adaptive high-resolution model that divides input images into smaller grid patches, each encoded at the native resolution of 336×336, and then merges them into a large feature map. By removing image padding, it adjusts seamlessly to any aspect ratio. To add global context and reduce stitching artifacts, it also concatenates features from a downsampled image with the merged map, allowing efficient support for any resolution. Since the final image tokens are merged into two image sequences, we restore each image sequence to a two-dimensional arrangement and then perform our AdaptPrune under a unified pruning budget.

\textbf{InternVL2.} InternVL2 \cite{internvl1_5} is a high-resolution model, employing a dynamic high-resolution strategy by dividing images into 448×448 tiles, with aspect ratios matched from a predefined set to retain natural proportions, it scales up to 40 tiles to accommodate 4K resolution. A 448×448 thumbnail of the entire image is included alongside the tiles to provide global context. Pixel shuffling reduces each 448×448 tile to 256 visual tokens, enabling efficient high-resolution processing with a reduced token count. Since multiple sub-images tiles are directly inputted into the model for processing, we restore each sub-image to a two-dimensional arrangement and then perform AdaptPrune under a unified pruning budget. In our AdaptPrune, each sub-image is processed individually. During AdaptPrune token selection, we consider scores across all tokens but ensure that the adaptive suppression process does not cross sub-image boundaries.

\section{Datasets}
To demonstrate the broad applicability of our approach, we use several representative datasets that cover a wide spectrum of task domains for evaluation, including Image Captioning, General Visual Question Answering, Text-based Visual Question Answering, and Multimodal Reasoning.

\textbf{Image Captioning.} In Image Captioning, models generate a descriptive sentence for each image. We use the Nocaps \cite{agrawal2019nocaps} and Flickr30K \cite{plummer2015flickr30k} datasets as benchmarks. 
Nocaps \cite{agrawal2019nocaps}  is the first large-scale benchmark dataset featuring 166,100 human-generated captions for 15,100 images, designed to evaluate image captioning on novel objects.
The Flickr30k \cite{plummer2015flickr30k} dataset is a standard benchmark for sentence-based image description, featuring 31,000 images, each accompanied by five reference sentences provided by human annotators.
We prompt the model with the phrase "Provide a one-sentence caption for the provided image." for both benchmarks and use the CIDEr score \cite{vedantam2015cider} for measurement metric.

\begin{table*}[t]
\setlength{\tabcolsep}{1.2mm}
\centering
\small

  \begin{tabular}{l|ccccc}
    \toprule
    Category (dev)   & LLaVA-1.5-7B \cite{llava1.5}  & FastV \cite{fastv} 90\% & AdaptPrune 90\% & FastV \cite{fastv} 75\% & AdaptPrune 75\% \\ 
    \midrule
    Action Recognition     & 90.7 & 85.2 & 88.9 & 87.0 & 90.7 \\
    Attribute Comparison   & 50.0 & 50.0 & 54.5 & 52.3 & 54.5 \\
    Attribute Recognition  & 79.7 & 68.9 & 75.7 & 77.0 & 79.7 \\
    Celebrity Recognition  & 79.8 & 76.8 & 74.7 & 78.8 & 78.8 \\
    Function Reasoning     & 75.9 & 72.2 & 70.9 & 75.9 & 70.9 \\
    Future Prediction      & 45.0 & 30.0 & 37.5 & 40.0 & 42.5 \\
    Identity Reasoning     & 93.3 & 86.7 & 95.6 & 95.6 & 97.8 \\
    Image Emotion          & 78.0 & 58.0 & 78.0 & 78.0 & 76.0 \\
    Image Quality          & 35.8 & 22.6 & 30.2 & 28.3 & 32.1 \\
    Image Scene            & 96.2 & 90.4 & 94.2 & 96.2 & 97.1 \\
    Image Style            & 77.4 & 73.6 & 71.7 & 77.4 & 77.4 \\
    Image Topic            & 83.3 & 80.6 & 83.3 & 83.3 & 83.3 \\
    Nature Relation        & 41.7 & 39.6 & 37.5 & 37.5 & 37.5 \\
    Object Localization    & 39.5 & 35.8 & 38.3 & 37.0 & 38.3 \\
    OCR                    & 59.0 & 59.0 & 59.0 & 59.0 & 59.0 \\
    Physical Property Reasoning      & 50.7 & 53.3 & 49.3 & 53.3 & 50.7 \\
    Physical Relation      & 33.3 & 41.7 & 41.7 & 41.7 & 41.7 \\
    Social Relation        & 88.4 & 53.5 & 55.8 & 72.1 & 69.8 \\
    Spatial Relationship   & 17.8 & 17.8 & 17.8 & 17.8 & 17.8 \\
    Structured Image-Text Understanding  & 26.9 & 30.8 & 26.9 & 28.2 & 26.9 \\ 
    \bottomrule
  \end{tabular}
  \caption{MMBench \cite{liu2025mmbench} finegrained comparison between FastV and AdaptPrune at 90\% and 75\% pruning ratios.
  }
  \label{tab:finegrained_mmbench}
\end{table*}

\textbf{General Visual Question Answering.} Visual Question Answering (VQA) requires the model to generate an answer for a given image-question pair based on the content of the image. We evaluate the model's performance on four commonly used and representative datasets: VQAv2 \cite{vqav2}, OK-VQA \cite{marino2019okvqa}, VizWiz \cite{gurari2018vizwiz}, and GQA \cite{hudson2019gqa}. 
VQAv2 \cite{vqav2} is a manually annotated open-ended question and answer dataset about images, requiring an understanding of visuals, language, and common sense to answer the questions.
OK-VQA \cite{marino2019okvqa} is a benchmark for knowledge-based visual question answering, featuring over 14,000 questions that require external knowledge to answer, designed to challenge current VQA models and encourage the integration of external data sources.
VizWiz \cite{gurari2018vizwiz} dataset consists of images and spoken questions from blind users, each with 10 crowdsourced answers, challenging models to predict answers or identify unanswerable questions.
GQA \cite{hudson2019gqa} is a visual question answering dataset featuring real images with balanced Q\&A pairs, scene graph annotations, and pre-extracted visual features from advanced detection models.
The model is prompted with the instruction: "Answer the question using a single word or phrase." for all benchmarks. We use exact match accuracy as the metric for measurement.

\textbf{Multimodal Reasoning.} Multimodal reasoning demands more sophisticated perception, knowledge, and reasoning skills than VQA, which makes it a more comprehensive task for evaluating the integrated capabilities of LVLMs. We assess these skills with the datasets MME, MMBench \cite{liu2025mmbench}, and POPE \cite{pope}. 
MMBench \cite{liu2025mmbench} is a systematically designed benchmark for evaluating LVLMs, offering a diverse set of multi-choice questions, rigorous quality control, and a unique CircularEval strategy to ensure accurate and holistic model assessments.
MME is a comprehensive evaluation benchmark for LVLMs, designed to assess their perception and cognition abilities across 14 subtasks, with manually crafted instruction-answer pairs to ensure a fair comparison and minimize data leakage.
POPE \cite{pope} is an innovative evaluation method designed to more effectively and flexibly assess object hallucination in large vision-language models by utilizing a polling-based query approach.
The model is prompted with the instruction: "Answer the question using a single word or phrase." for MME. 
The model is prompted with the instruction: "Answer with the option's letter from the given choices directly." for MBench \cite{liu2025mmbench}. 
We use accuracy as the metric for MME and POPE \cite{pope} and GPT Evaluation Score by GPT-3.5-Turbo-0613 as the metric for MMBench \cite{liu2025mmbench}.

\begin{table*}[t]
\setlength{\tabcolsep}{1.2mm}
\centering
\small

  \begin{tabular}{l|ccccc}
    \toprule
    L2-category (dev)  & LLaVA-1.5-7B \cite{llava1.5} & FastV \cite{fastv} 90\% & AdaptPrune 90\% & FastV \cite{fastv} 75\% & AdaptPrune 75\% \\
    \midrule
    Attribute Reasoning & 70.4 & 68.3 & 68.3 & 71.9 & 69.3 \\
    Coarse Perception & 77.4 & 68.6 & 74.7 & 76.0 & 76.7 \\
    Finegrained Perception (Cross-Instance) & 55.2 & 53.1 & 55.9 & 54.5 & 56.6 \\
    Finegrained Perception (Instance-Level) & 65.9 & 61.1 & 62.8 & 64.2 & 65.2 \\
    Logic Reasoning & 33.1 & 30.5 & 30.5 & 32.2 & 32.2 \\
    Relation Reasoning & 57.4 & 45.2 & 45.2 & 51.3 & 50.4 \\
    \bottomrule
  \end{tabular}
  \caption{MMBench \cite{liu2025mmbench} L2-category finegrained comparison between FastV and AdaptPrune at 90\% and 75\% pruning ratios.}
  \label{tab:finegrained_mmbench_l2}
\end{table*}

\begin{table*}[t]
\setlength{\tabcolsep}{1.2mm}
\centering
\small

  \begin{tabular}{l|ccccc}
    \toprule
    Category & LLaVA-1.5-7B \cite{llava1.5} & FastV \cite{fastv} 90\% & AdaptPrune 90\% & FastV \cite{fastv} 75\% & AdaptPrune 75\% \\ 
    \midrule
    Code Reasoning & 67.5 & 45.0 & 62.5 & 70.0 & 52.5 \\
    Numerical Calculation & 70.0 & 47.5 & 57.5 & 52.5 & 55.0 \\
    Text Translation & 107.5 & 125.0 & 117.5 & 120.0 & 115.0 \\
    Commonsense Reasoning & 112.9 & 111.4 & 112.9 & 120.0 & 117.1 \\
    Artwork & 119.5 & 106.5 & 113.8 & 120.0 & 117.5 \\
    Celebrity & 137.1 & 127.4 & 133.2 & 138.5 & 134.7 \\
    Count & 155.0 & 95.0 & 138.3 & 143.3 & 145.0 \\
    Color & 170.0 & 130.0 & 155.0 & 170.0 & 170.0 \\
    Position & 128.3 & 86.7 & 115.0 & 130.0 & 126.7 \\
    OCR & 140.0 & 125.0 & 125.0 & 147.5 & 132.5 \\
    Landmark & 163.8 & 136.5 & 154.0 & 157.5 & 166.0 \\
    Scene & 158.0 & 147.5 & 165.8 & 159.8 & 163.3 \\
    Existence & 190.0 & 135.0 & 165.0 & 160.0 & 170.0 \\
    Posters & 146.6 & 128.2 & 144.9 & 142.5 & 149.3 \\
    \bottomrule
  \end{tabular}
  \caption{MME finegrained comparison between FastV and AdaptPrune at 90\% and 75\% pruning ratios.}
  \label{tab:finegrained_mme}
\end{table*}

\textbf{Text-based VQA.} Text-based visual question answering focuses on understanding textual information within images. We evaluate this capability with the datasets TextVQA \cite{textvqa}, DocVQA \cite{docvqa}, and ChartQA \cite{masry2022chartqa}.
The TextVQA \cite{textvqa} dataset focuses on visual question answering in scene text scenarios, challenging models to read and interpret text within images to answer related questions.
The DocVQA \cite{docvqa} dataset offers 50,000 questions across over 12,000 document images, challenging models to understand document structure and content for Visual Question Answering.
ChartQA \cite{masry2022chartqa} is a large-scale benchmark with over 32.7K questions, combining human-written and generated queries from chart summaries, designed to test complex visual and logical reasoning on charts.
The model is prompted with the instruction: "Answer the question using a single word or phrase." for all benchmarks.
We use exact match accuracy as the metric for TextVQA \cite{textvqa} and ChartQA \cite{masry2022chartqa} and ANLS Score as the metric for DocVQA \cite{docvqa}.

\section{Pruning Strategy Ablation}

We conduct a series of ablation studies on our pruning strategy with comparison methods commonly used in other works \cite{fastv,fitprune,chen2024llavolta}. Our experiments include: (a) a single-layer implementation of FitPrune, which uses the cross attention between image and text multiplied by the self attention among image tokens; (b) selecting every other token from the top 20\% tokens; (c) randomly selecting 10\% tokens from all image tokens; (d) randomly selecting one token from each 3x3 grid and then choosing the top 10\% tokens; (e) selecting the highest scoring token from each 3x3 grid, and then choosing the top 10\% tokens; (f) applying average pooling with a stride of 1 and a kernel size of 3 over the two-dimensional image; (g) selecting the last 10\% of the image token sequence. The experiment results demonstrate that our method is the most effective.

\section{Finegrained Results}

\subsection{MMBench Finegrained Results}
As shown in Table \ref{tab:finegrained_mmbench}, in the MMBench \cite{liu2025mmbench} fine-grained comparison between FastV \cite{fastv} and AdaptPrune at 90\% and 75\% pruning ratios, significant performance improvements are evident with AdaptPrune in several categories. Specifically, AdaptPrune shows enhanced outcomes in Action Recognition, Attribute Recognition, Future Prediction, Identity Reasoning, Image Emotion, Image Quality, and Image Scene. These results underline AdaptPrune's ability to retain crucial visual information for complex understanding and response capabilities within dynamic environments.

The MMBench \cite{liu2025mmbench} L2-category comparison in Table \ref{tab:finegrained_mmbench_l2} reveals that AdaptPrune generally improves performance in categories such as Coarse and Finegrained Perception. This indicates that AdaptPrune’s nuanced pruning approach could be beneficial for detailed perceptual analysis.

\subsection{MME Finegrained Results}
In the MME comparison in Table \ref{tab:finegrained_mme}, AdaptPrune's strategy demonstrates considerable success in categories including Code Reasoning, Commonsense Reasoning, Artwork, Count, Color, Position, Landmark, Scene, Existence, and Posters. These improvements suggest that AdaptPrune is particularly effective in contexts requiring an intricate understanding of visual and contextual elements, arguably due to its capability to preserve essential tokens that encapsulate critical information for interpreting complex scenarios.

\section{Visual Attention Distribution}
In Figure \ref{fig:supplement_attention_layer} and Figure \ref{fig:supplement_attention_layer_vqav2}, we present the average attention score distribution across each layer of LLaVA-1.5-7B. We randomly choose 1,000 samples from the VQAv2 \cite{vqav2} benchmark of the General VQA task and the MMBench \cite{liu2025mmbench} benchmark of the Multimodal Reasoning task for analysis. 
The model exhibits an almost identical attention distribution pattern for image tokens within the same layer, even though these samples come from entirely different tasks.
The pattern changes slightly across different layers, the pattern changes slightly, but some of the tokens receiving the highest attention remain consistent.
The detailed statistical results further reveal that the model's attention scores exhibit specific positional bias.

\begin{figure*}[t]
  \centering
  \includegraphics[width=.9\linewidth]{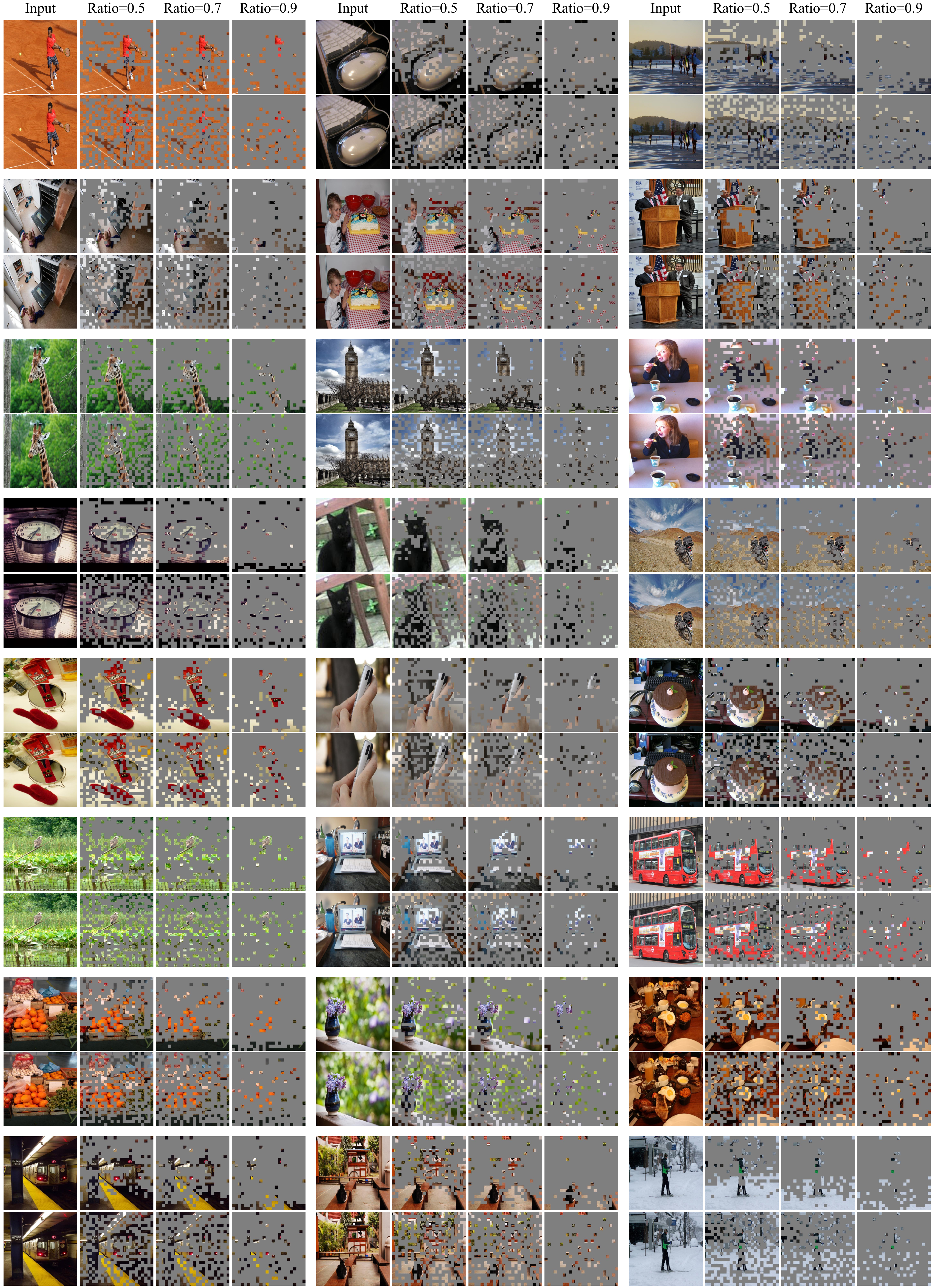}
  
  \caption{Uncurated random sample comparison on VQAv2 \cite{vqav2} benchmark images. For each comparison group with 8 images, the top row shows FastV \cite{fastv} results and the bottom row shows AdaptPrune results, with columns from left to right for input images and pruning ratios of 0.5, 0.7, and 0.9.
  }
  \label{fig:supplement_ratio}
\end{figure*}

\begin{figure*}[h]
  \centering
  \includegraphics[width=.95\linewidth]{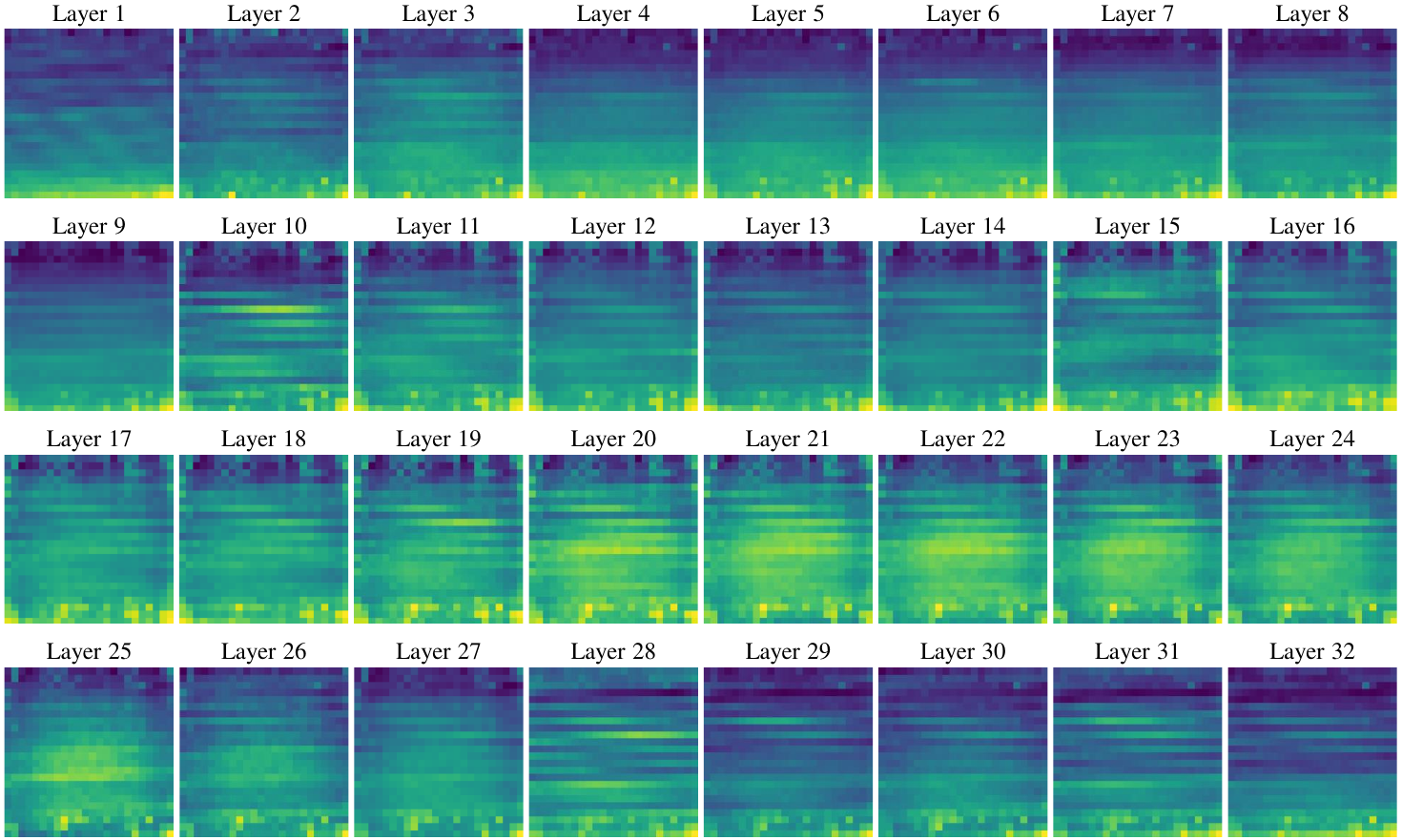}
  
  \caption{
  Visualization of the average attention distribution across each layer of LLaVA-1.5-7B. The attention scores are computed from 1,000 random samples of VQAv2 \cite{vqav2}.
  }
  \label{fig:supplement_attention_layer}
\end{figure*}

\begin{figure*}[h]
  \centering
  \includegraphics[width=.95\linewidth]{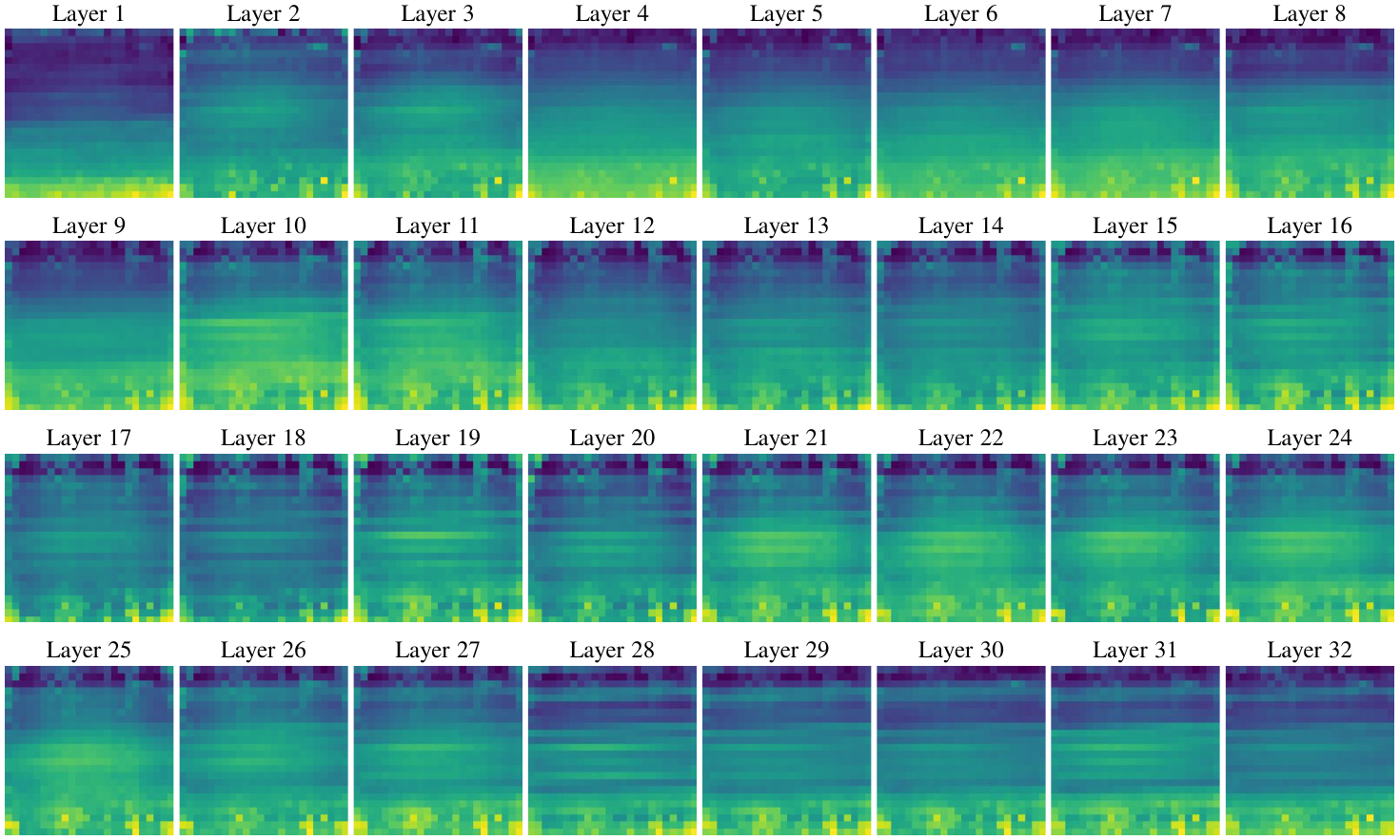}
  
  \caption{
  Visualization of the average attention distribution across each layer of LLaVA-1.5-7B. The attention scores are computed from 1,000 random samples of MMBench \cite{liu2025mmbench}.
  }
  \label{fig:supplement_attention_layer_vqav2}
\end{figure*}

\end{document}